\documentclass[10pt,twocolumn,letterpaper]{article}

\usepackage[pagenumbers]{cvpr}
\usepackage{multirow}
\usepackage{bm}
\usepackage{amssymb}
\usepackage{amsmath}
\usepackage{array}   
\usepackage{tabularx}
\usepackage{multirow}
\usepackage{amsmath, amsfonts, amsthm, stackrel, amssymb}   
\usepackage{marvosym}
\definecolor{cvprblue}{rgb}{0.21,0.49,0.74}
\usepackage[pagebackref,breaklinks,colorlinks,allcolors=cvprblue]{hyperref}

\def\paperID{2960} 
\def\confName{CVPR}
\def\confYear{2025}

\title{Bidirectional Sparse Attention for Faster Video Diffusion Training}

\begin{document}
\author{
    Chenlu Zhan$^{1*}$\quad 
    Wen Li$^{1}$\thanks{Equal Contribution.} \quad 
    Chuyu Shen$^{1}$\quad 
    Jun Zhang$^{1}$\quad 
    Suhui Wu$^{1}$\quad 
    Hao Zhang$^{1\hat{}}$\thanks{Project Leader. $^{\hat{}}$Corresponding Author.} \\
    $^1$ByteDance \\
    \small \texttt{zhanchenlu@bytedance.com, liwen.8459@bytedance.com, zhanghao.25@bytedance.com}
}

\maketitle
\begin{abstract}
Video diffusion Transformer (DiT) models excel in generative quality but hit major computational bottlenecks when producing high-resolution, long-duration videos. 
 Full attention requires computing dot products between all pairs of queries and keys, resulting in a quadratic computational complexity with respect to the sequence length (\(O(L^2)\)), leading to high training and inference costs.
To overcome this limitation, we propose a Bidirectional Sparse Attention (BSA) framework that sparsification from both the query and key–value directions to reduce the quadratic cost of full attention. 
Specifically, the sparsification of queries is achieved by pruning tokens that are locally redundant or semantically similar within the 3D spatiotemporal domain of a video. For the key–value pairs, only those that are highly correlated with the query at a global level are selected for attention computation. Furthermore, we design a dynamic threshold adjustment mechanism to adaptively regulate the selected number of key–value pairs, thereby maintaining quality without degradation.
Extensive experiments demonstrate that BSA significantly accelerates DiT training across long sequences, reducing FLOPs by up to \textbf{20×} and achieving \textbf{17.79×} faster attention training, while preserving or even surpassing the generative quality of full attention.
\end{abstract}

\section{Introduction}
\label{sec:intro}
Modeling long sequences remains a pivotal challenge in deep learning, particularly for video diffusion models (VDMs) designed to generate long-duration high-resolution videos~\cite{rombach2022high,peebles2023scalable,vaswani2017attention}. 
Although full attention is powerful, its quadratic complexity with respect to sequence length ($O(L^2)$) 
severely limits scalability. 
\begin{figure}[h]
\centering
\includegraphics[width=0.9\linewidth]{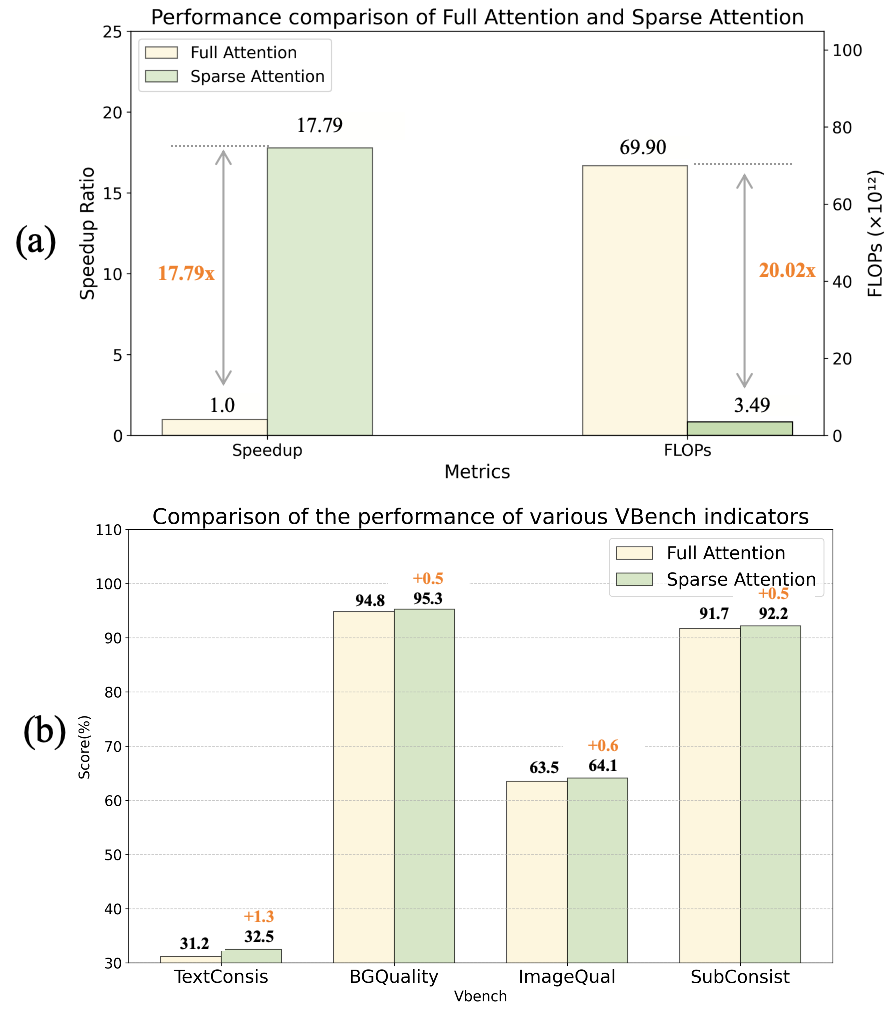}
\caption{Performance comparison between our sparse attention with full attention. (a) Speedup ratio and computational cost. (b) Generation quality across four consistency metrics on VBench~\cite{huang2024vbench}.}
\label{fig:0}
\end{figure}
Even a brief few-second video clip can expand into hundreds of thousands of tokens, making attention computation the primary bottleneck in video Diffusion Transformers (DiTs)~\cite{wan2025wan,kong2024hunyuanvideo,ma2025step,yang2024cogvideox}. Moreover, attention operations account for over 90\% of training costs in VDMs, highlighting the critical need for efficient and trainable sparse attention mechanisms.

\begin{figure}[h]
\centering
\includegraphics[width=1\linewidth]{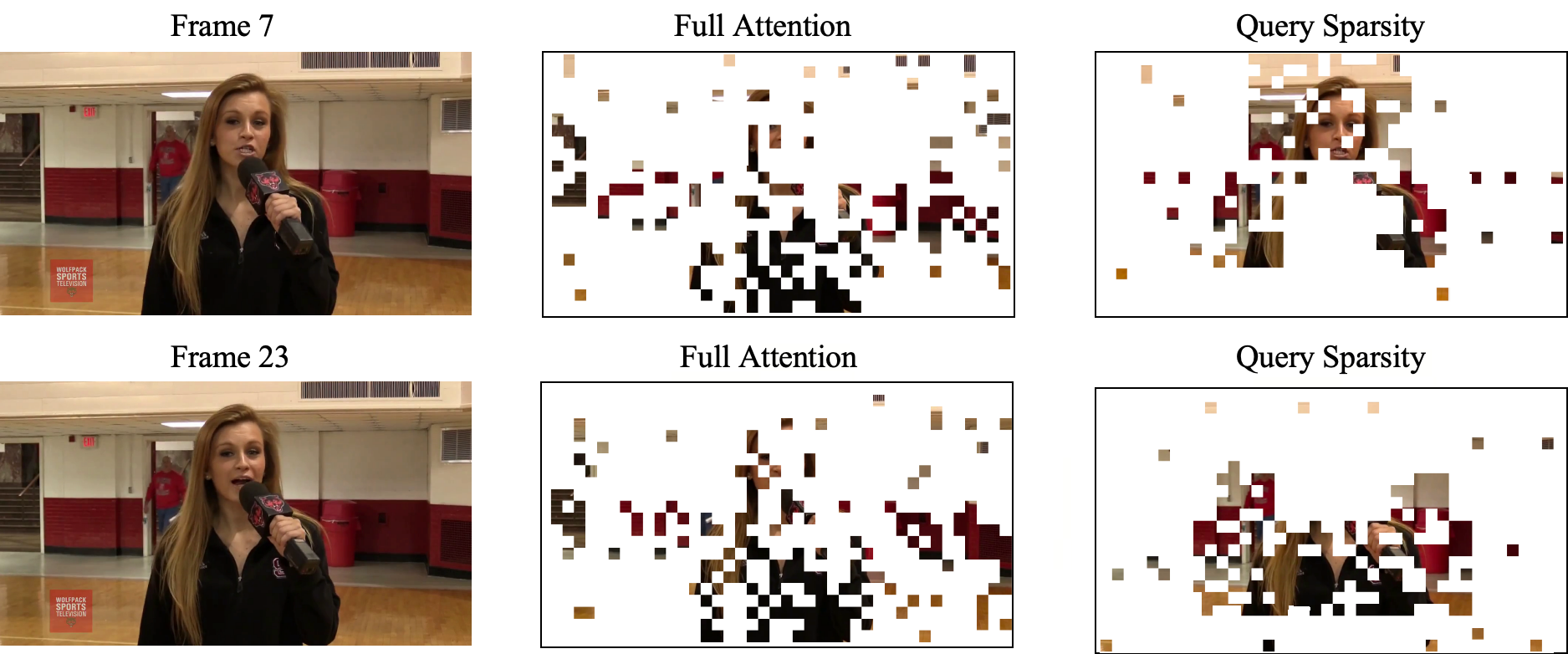}
\caption{Distribution of important queries in attention computation. In full attention, the attention maps across different frames are highly similar, indicating the presence of redundant queries that lead to repetitive computations. In contrast, our Query-Sparse attention produces distinct attention maps that focus on salient content such as human actions, rather than static backgrounds. This demonstrates the method’s ability to prune redundant features while preserving essential semantics.
}
\label{semantic}
\end{figure}

To design an efficient trainable sparse attention mechanism, we first pinpoint two primary sources of training latency:
\textbf{(1) Bilateral Redundancy.} Previous studies\cite{lu2025moba,yuan2025native,zhang2025vsa} have shown that there is redundancy in key–value representations during full attention and that only a small subset of key–value pairs contributes meaningfully to each query. In our observation, the same redundancy also applies to queries. As shown in Fig.~\ref{semantic}, we identify the most important queries in the attention computation based on their similarity to the keys. Specifically, all queries are ranked according to their cumulative similarity scores and retain only the top 20\%. We visualize two similar frames from a video and observe that in full attention, the distributions of important queries in these two frames are highly similar, indicating redundancy and repetitive computation between them. In contrast, our proposed Sparse-Query attention prunes less informative queries and retains only a subset for computation. The remaining queries in the two frames focus on different regions, primarily the moving foreground areas, and complementing each other. This shows that pruning query tokens not only reduces computational cost but also enhances the model’s focus on semantically important regions.
\textbf{(2) Dynamic Sparsity.} Each query's matching critical subset of KV pairs dynamically changes~\cite{wu2025vmoba}. Besides, the sparsity of different local KV subsets varies dynamically at different training steps~\cite{tan2025dsv}.
Building upon these observations, recent works~\cite{lu2025moba,zhang2025vsa,yuan2025native,tan2025dsv} introduce sparse attention by restricting each query to a fixed subset of KV pairs~\cite{lu2025moba,zhang2025vsa}. However, they only prune KV pairs, ignoring query-side redundancies like repeated semantics across frames, which further inflate costs. Moreover, fixed sparsity patterns are parameter-sensitive, and over or under sizing can cause computation redundancy and performance loss, failing to adapt to real dynamic sparsity. Overall, these sparse methods have not fully adapted to the sparsity discoveries in attention, remaining limited.

To address the above challenges, we propose \textbf{BSA}, a  \textbf{B}idirectional  \textbf{S}parse \textbf{A}ttention, a novel framework designed to accelerate video diffusion training and inference.
Our primary contribution is the first explicit analysis that leverages the inherent sparsity in Query, combining it with KV sparsity to formulate a bidirectional sparse-attention mechanism. This approach is further enhanced by a dynamic adjustment strategy that achieves adaptive sparsity tuning.
Specifically, our method differs in two major aspects:
(1) \textbf{Query Sparsity.}  We begin by partitioning long-sequence tokens into 3D blocks.
For query sparsity, we preserve tokens with distinctive information and prune redundant ones based on their similarity to the block's center token.
(2) \textbf{Dynamic Statistic KV Sparsity.}  
    For Sparse KV, we dynamically identify the most relevant KV tokens for each query. Using block-wise attention scores, we compute statistic thresholds and iteratively admit tokens until a cumulative score target is reached.
Our approach is model-agnostic and can be applied to any DiT architecture. 
We implement BSA on Wan2.1-1.3B and Wan-2.1-14B model with from-scratch training. 
Extensive experiments demonstrate that BSA significantly accelerates training, as shown in Fig.~\ref{fig:0},
achieving up to \textbf{20$\times$} reduction in FLOPs and \textbf{17.79$\times$} faster attention computation, 
while preserving or even surpassing the generative quality of full attention.  
Furthermore, our method can be seamlessly applied training-free, achieving a \textbf{6.2$\times$} end-to-end speedup on an H100 GPU while maintaining superior generation quality.
Our contributions are summarized as follows: 
\begin{itemize}
    \item We present BSA, a trainable bidirectional dynamic sparse-attention framework, which for the first time orthogonally sparsifies Query and Key–Value pairs to accelerate video diffusion training and inference.

    \item We devise distinct dynamic sparsity strategies for Queries and KV blocks, effectively capturing attention variability during training and enabling adaptive token selection beyond fixed patterns.
    \item Extensive experiments on Wan2.1-1.3B show that BSA delivers up to a 20x FLOPs reduction and substantial acceleration, 17.7x in training, 6x in inference, and 6.2x in training-free applications, all while maintaining or surpassing the generative quality of full attention.
\end{itemize}
\section{Related Work}
\label{sec:intro}
\noindent\textbf{General Sparse Attention.}
Sparse attention has been widely adopted in Large Language Models (LLMs)~\cite{guo2021longt5,ding2023longnet,beltagy2020longformer,lou2024sparser,gao2024seerattention,gao2025seerattention,lai2025flexprefill,yang2025lserve,wang2024qsparse} and Vision–Language Models (VLMs)~\cite{shen2025fastvid,zhang2025spargeattentionaccuratetrainingfreesparse,li2025mminference,qin2025video} to mitigate the computational explosion caused by long input sequences.
However, most methods fix sparsity a prior. For instance, LLM schemes~\cite{xiao2023efficient,han2023lm} exploit attention concentration on early or local tokens. SeerAttentions~\cite{gao2024seerattention,gao2025seerattention} chiefly prune token redundancy under causal masks. MInference~\cite{jiang2024minference} leverages head-wise heterogeneous sparsity.
However, these techniques are limited to predefined patterns, neglecting the inherent dynamic redundancy of video data, and are typically optimized for inference acceleration rather than training. 
To tackle this, recent methods like MoBA~\cite{lu2025moba} and NSA~\cite{yuan2025native} explore trainable dynamic sparsity for end-to-end acceleration on long sequences. However, these works~\cite{liu2024clearconvlikelinearizationrevs,zhang2025flashvideo,zhang2025sageattention3} concentrate solely on redundancy in Key-Value pairs, disregarding dynamic redundancy in Queries. Besides, they frequently depend on fixed selection rules that fail to adapt to the inherently dynamic sparsity of sequences, leading to additional computational overhead.

\begin{figure*}[h]
\centering
\includegraphics[width=1\linewidth]{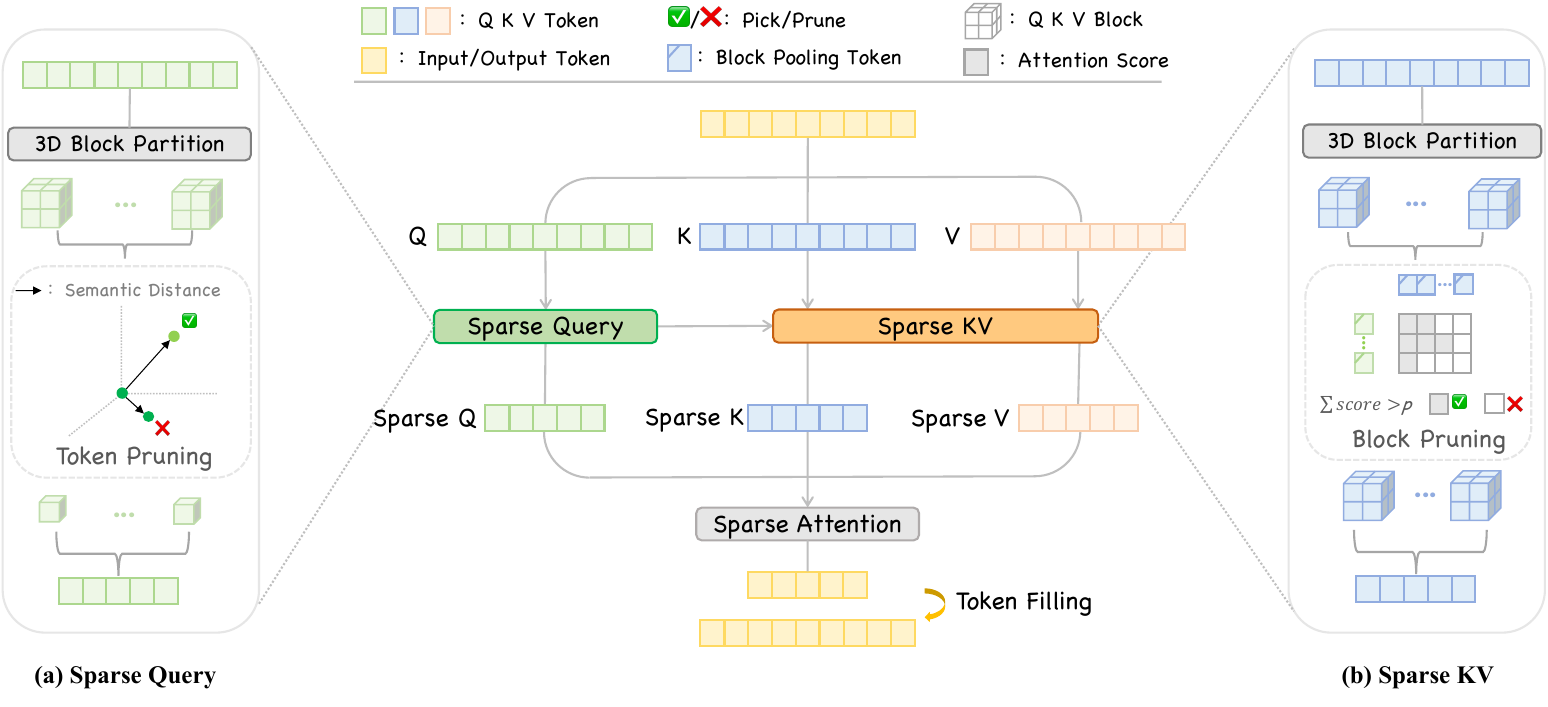}
\caption{{Overview of  BSA.} 
We introduce a Bidirectional Attention Sparsification that exploits the dynamic sparsity of both Queries and Key-Value (KV) pairs. The $QKV$ sequences of video are first partitioned into blocks to efficiently select critical tokens (Sect.~\ref{block_partition}). (a) We then select each query block’s center token, linearly score within-block tokens by semantic similarity to the center, and prune a fixed fraction of redundant tokens to retain only the most informative queries (Sec.~\ref{Sparse Query}). (b) For KV sparsity, we dynamically pick the most relevant KV blocks for each query block and prune the unrelated KV blocks. We compute sparsity-adaptive thresholds and iteratively admit tokens until a cumulative score target is met (Sec.~\ref{Sparse KV}). }
\label{main}
\end{figure*}
\noindent\textbf{Sparse Attention for Video Diffusion.}
Recent works~\cite{tan2025dsv,xi2025sparse,lou2024sparser,ding2025efficient,hassani2025generalized,yang2025sparse,yuan2024ditfastattn,cai2025mixture,wang2025lavie,ma2025latte,li2025radial,ren2025grouping} transfer sparse attention from LLMs to video DiTs~\cite{zhang2025magicmirror,kong2024hunyuanvideo,wan2025wan,yang2024cogvideox,zhu2025scaling,xing2025motioncanvas}, mainly to accelerate inference. However, training-free schemes~\cite{xia2025training} are unsuitable for video pre-training: fixed sparsity patterns rely on fragile hyperparameters~\cite{lu2025moba,zhang2025vsa}, eroding quality and causing artifacts in long video sequences~\cite{xi2025sparse,yuan2024ditfastattn,zhang2025training}; additionally, quadratic attention over long sequences bottlenecks both training and inference. Thus, a trainable sparse-attention mechanism is needed. While VSA~\cite{zhang2025vsa} (fixed block sizes, top-k selection) and VMoBA~\cite{wu2025vmoba} (thresholding, sensitive to design) offer partial relief, they overlook query-side redundancy and fail to track video dynamics, leading to wasteful computation. To address this, we propose a bidirectional sparse-attention scheme that dynamically prunes both Queries and Key-Value pairs, eliminating redundancy while preserving generative fidelity.


\section{Method}
In this section, we introduce the \textbf{BSA}, a \textbf{B}idirectional \textbf{A}ttention \textbf{S}parsification method that exploits the sparsification of both Queries and Key-Value (KV) pairs. 
Our method removes query-side redundancy and adaptively selects salient KV pairs, yielding near-optimal bidirectional optimization under quadratic attention. Specifically, Sec.~\ref{review} surveys the key components and motivation of sparse attention. Sec.~\ref{block_partition} presents block partitioning, and Sec.~\ref{Sparse Query} details the query-side sparsification. We describe the dynamic statistic KV-sprarse in Sec.~\ref{Sparse KV}. We also introduce the specific kernel design in Sec.~\ref{kernal}.

\subsection{Review of Sparse Attention}
\label{review}

Modern video DiTs use 3D full attention to model dependencies across the entire spatiotemporal volume. Given a latent tensor $(T, H, W)$, each token at $(t, h, w)$ is flattened to a 1D index $n = t H W + h W + w$, yielding a sequence of length $L = THW$. Full attention is then applied to this sequence, enabling all-to-all token interactions.
For a single attention head, let $Q, K, V \in \mathbb{R}^{L \times d}$ denote the query, key, and value matrices, 
and let $M \in \{-\infty, 0\}^{L \times L}$ denote the attention mask that specifies the allowed token-to-token connections. 
The attention output $O$ is computed as:
   \begin{equation}
S = \frac{Q K^{\top}_s}{\sqrt{d_k}}, \quad 
O = \text{Softmax}(S)V_s .
\end{equation}

In full attention, all sequence tokens from Q, K, and V participate in interaction and computation. Sparse attention, which theoretically reduces overall computation by selecting a key subset $K_s$ and $V_s$ from the KV pairs, aims to improve efficiency.

\subsection{BSA Structure}

As illustrated in Fig.~\ref{main}, our framework consists of three main components:  
(a) \textbf{3D Block Partition.} 
The video latent is divided into spatiotemporal blocks to efficiently filter critical information.  
(b) \textbf{Sparse Query.}  
We efficiently select the most informative query tokens while pruning the redundant ones.  
(c) \textbf{Sparse KV.}  
The most relevant KV tokens are selected for each query. The selection strategy is dynamic, iteratively accruing candidates until the cumulative score meets a target threshold. This eliminates fixed sparsity patterns and adapts naturally to content-dependent variability.

\subsubsection{Block Partition}
\label{block_partition}
Due to the fact that existing general-purpose acceleration techniques rely on structured processing of the sequence, such as $Flash Attention$, which partitions the sequence into GPU-friendly blocks for parallel computation, our token sparsification also operate on block-partitioned sequences to translate theoretical sparsity into real speedup. Given a video latent of shape $(T, H, W)$, it is partitioned into blocks of size $B = (C_t, C_h, C_w)$, where each block corresponds to a GPU block. The total number of blocks is thus $(N_t, N_h, N_w) = (T / C_t, H / C_h, W / C_w)$. Besides, video latent exhibits inherent spatiotemporal structure, which can be disrupted when the 3D sequence is flattened into a 1D sequence for attention computation. To preserve this structure, we record the original 3D index of each token and ensure that, after block partitioning, tokens that are adjacent in the video space remain adjacent within the sequence.




\subsubsection{Sparse Query}
\label{Sparse Query}

 
Video data intrinsically contains temporal dependencies across frames and spatial coherence within frames, leading to substantial spatiotemporal redundancy. As illustrated in Fig. \ref{semantic}, full attention contains a substantial amount of redundant query tokens, while core information is primarily carried by only a small subset of important queries. We address this by proposing a feature-redundancy-based query-sparse method to eliminate redundancy.
Given a query sequence $Q$ partitioned into $N$ blocks $Q_c$, let $Q_c^{(b)}$ denote the tokens in block $b$ and $q_c^{(b)}$ its critical token (central token usually). Tokens within a block (e.g., spatial neighbors) typically share similar features, allowing $q_c^{(b)}$ to represent the block. We thus measure the cosine similarity between each token and $q_c^{(b)}$ within $Q_c$, preserving local distinctions and avoiding the uniformity of average pooling.

For each block, token pruning is implemented on tokens with higher similarity, and only non-redundant tokens with lower similarity are retained. These selected tokens drive core attention scores, while redundant ones, whose features are largely duplicative, are safely pruned without loss of performance or semantic integrity.
For each block, a pruning ratio $r$ is applied to retain a portion of tokens. The retained tokens from all blocks are concatenated to form a new redundancy-free query sequence $Q^s$, defined as:
   \begin{equation}
   \resizebox{0.9\hsize}{!}{$
Q^{s} = \bigcup_{b=1}^{N} \left\{ q_i \in Q_c^{(b)} \;\middle|\; 
\operatorname{rank}_b\big(1 - \cos(q_c^{(b)}, q_i)\big) \leq \lceil r \cdot |Q_c^{(b)}| \rceil \right\}$}
\end{equation}
where $\operatorname{rank}_b(\cdot)$ is the ranking within block in descending order, $N$ is the number of blocks. 
$|Q_c^{(b)}|$ is the number of tokens in block $b$, and $r \in (0,1]$ is the retained ratio.  

\noindent\textbf{Token Filling.} After pruning the query sequence, the output sequence produced by the attention operation no longer matches the original sequence length, since only the retained queries participate in the computation. To address this mismatch, we restore the original length through a token-filling procedure. During query pruning, the discarded tokens are those that are highly similar to their corresponding center token within each block. Therefore, after the center token undergoes attention computation, its output can serve as a reliable approximation for the outputs of the pruned tokens. By filling the missing positions with the center token’s output, we reconstruct a sequence with the same length as the original input.

\noindent\textbf{Critical Token.} To enhance critical token selection, we adopt a window-based mechanism: each block query is further partitioned into smaller windows of size $(w_t, w_h, w_w)$, evenly dividing the block $(C_t, C_h, C_w)$. Within each window, the center token is chosen and its similarity to neighboring tokens is computed via cosine similarity. Tokens selected from all windows are then concatenated, effectively preserving non-redundant, semantically important tokens.

\subsubsection{Sparse KV}
\label{Sparse KV}
For sparsification of KV, the most relevant KV are selected for each query by the inter-block attention score. Building on block-level representations from cuboid partitioning, each query interacts with only a subset of KV pairs, greatly reducing computational cost. The core challenge is determining the appropriate subset size. Different queries require different numbers of KV subsets. However, a fixed strategy is inherently limited: over-selection leads to computational redundancy, whereas under-selection impairs performance. Thus, a fixed strategy fails to adaptively assign KV pairs. 
To address this, we propose a \textit{dynamic Sparse KV} method based on statistical thresholds, which adaptively selects the relevant KV pairs for each query. The sparsity threshold is determined from the input attention scores, eliminating the need for pre-defined sparse patterns and enabling adaptation to diverse input content.
Our dynamic sparsity manifests in two aspects:

\noindent\textbf{Computation of statistical dynamic threshold.} Specifically, given inter-block attention scores $s_b$ for a computation, we aim to select $k$ core KV pairs. Since the distribution of $s_b$ varies across inputs, a fixed $k$ may fail to capture truly critical KV pairs. We therefore compute a dynamic threshold $p$ based on the mean and standard deviation of the attention scores, such that $k$ key samples are selected:
   \begin{equation}
p = \text{mean}(S_b) + \text{std}(S_b) \cdot U(1 - k/n),
\end{equation}
where $n$ is the number of inter-block attention scores, $1 \le k \le n$ is the number of desired key samples, and $U(\cdot)$ denotes the quantile function.

\noindent\textbf{Dynamic selection of key KV pairs per query.} For the selected key blocks (assume $K$ in total), each query block computes attention with the KV pairs and dynamically selects indices. For query block $i$, the minimal index set $S_i$ is chosen such that the cumulative attention satisfies:
   \begin{equation}
      \resizebox{0.9\hsize}{!}{$
\gamma\big(\min |S_i| \quad \text{s.t.} \quad \sum_{(i,j)\in S_i} \frac{\exp(Q_i K_j^\top)}{\sum_{j'} \exp(Q_i K_{j'}^\top)} \ge p \big)$}
\end{equation}
while $\gamma$ denotes the operation that returns the minimal index set $S_i$, $p$ is the dynamic threshold. The results from all query blocks are then aggregated.

\subsubsection{Computation Cost Analysis} 
Let the sparsified query matrix be $Q^s \in \mathbb{R}^{rL \times d}$, where $rL$ is the number of sparsified query tokens, and the selected key and value matrices are $K_S$ and $V_S$, defined as
$K_S = \{ K_j \mid j \in \bigcup_i S_i \}, 
V_S = \{ V_j \mid j \in \bigcup_i S_i \}.$.
The sparse mask $M_S$ ensures attention is computed only between selected queries and KV pairs. The sparse attention output $O^s$ is expressed as:
   \begin{equation}
S^s = \frac{Q^s K_S^\top}{\sqrt{d_k}}, \quad
O^s = \text{Softmax}(S^s) V_S
\end{equation}
where $S^s \in \mathbb{R}^{rL \times L_S}$ is the sparsification attention score matrix ($L_S$ is the number of selected KV tokens), $A^s$ is the normalized sparse attention weight matrix, $\sqrt{d_k}$ is the scaling factor, and $O^s$ has the same sequence length as the input. 
We also provide a detailed analysis of the additional computational overhead introduced by our proposed query-sparse and KV-sparse methods. For Sparse Query, we compute intra-block token similarity and perform sorting over the sequence, resulting in a total computational complexity of $O(L)+O(LlogB)$, where $L$ is the sequence length and $B$ is the block size. For Sparse KV, similarity computation and sorting are performed between each query block and KV block, with a total complexity of $O(N)+O(N)$, where $N$ is the number of blocks. It is important to emphasize that the additional computation introduced by query and KV sparsification accounts for less than 0.1\% of the total FLOPs, making it a negligible cost.

\subsection{Kernal Design}
\label{kernal}
We implement separate forward and backward kernels with Triton, enabling hardware-efficient block-sparse attention for Flash Attention-level acceleration in training and pre-filling. By partitioning the attention mask into blocks, each GPU SM tile can process or skip blocks entirely, maximizing hardware efficiency where dense computation is preferred.
Query-sparse design creates block-level sparsity with variable block sizes. We use tailored kernels and mapping indices to compute only relevant pairs, maintaining dimension consistency. In both forward and backward passes, sparsity masks and mappings align blocks and filter excess tokens, ensuring proper computation throughout.

Our KV-sparse strategy adaptively selects variable-sized key KV pairs for each query, using custom kernels for efficient computation. Each $Q$ block records its attended KV blocks with $q2k\_num$ and $q2k\_index$, enabling distinct selections. Attention weights are computed and aggregated block-wise, with all outputs concatenated to form the final attention result. More details are in the Appendix.

\begin{table*}
	\centering
	\scalebox{0.7}{
\small
	\setlength{\tabcolsep}{8pt}
	
\begin{tabular}{llccccccc}
\hline
\multicolumn{1}{c}{\multirow{2}{*}{Seq\_len}}                                         & \multicolumn{1}{c}{\multirow{2}{*}{Method}} & \multirow{2}{*}{Sparsity} & \multicolumn{4}{c}{Quality}                                               & \multicolumn{2}{c}{Efficiency}          \\ \cline{4-9} 
\multicolumn{1}{c}{}                                                                  & \multicolumn{1}{c}{}                        &                           & TextConsis ↑     & BGConsis ↑       & ImageQual ↑      & SubConsist ↑     & ↓ FLOPs                & SpeedUp↑       \\ \hline
\multirow{2}{*}{\begin{tabular}[c]{@{}l@{}}61*448*832\\ 23,296 tokens\end{tabular}}    & Full Attention                              & -                         & 32.71\%          & 95.12\%          & \textbf{64.33}\%          & 92.34\%          & $1.51 \times 10^{12}$           & -              \\
                                                                                      & \textbf{Sparse Attention (Ours)}            & \textbf{0.93}              & \textbf{32.79\%} & \textbf{95.22\%} & {64.29\%} & \textbf{92.39\%} & $1.05 \times 10^{11}$             & \textbf{12.85x} \\ \hline
\multirow{2}{*}{\begin{tabular}[c]{@{}l@{}}157*768*1280\\ 153,600 tokens\end{tabular}} & Full Attention                              & -                         & 34.76\%          & 93.26\%          & 65.91\%          & 93.79\%          & $6.99 \times 10^{13}$          & -              \\
                                                                                      & \textbf{Sparse Attention (Ours)}            & \textbf{0.95}             & \textbf{34.93\%} & \textbf{93.41\%} & \textbf{66.03\%} & \textbf{94.13\%} &  $3.49 \times 10^{12}$  & \textbf{17.79x} \\ \hline
\end{tabular}}
\caption{Performance of the training-based BSA. We compare the generation quality and efficiency between BSA and full attention, based on the Wan2.1-1.3B model.} 
	\label{table1}
\end{table*}

\begin{table*}[!htp]
	\centering
	\scalebox{0.7}{
\small
	\setlength{\tabcolsep}{10pt}
	
\begin{tabular}{llccccccc}
\hline
\multicolumn{1}{c}{\multirow{2}{*}{Seq\_len}}                                         & \multicolumn{1}{c}{\multirow{2}{*}{Method}} & \multirow{2}{*}{Sparsity} & \multicolumn{4}{c}{Quality}                                               & \multicolumn{2}{c}{Efficiency}         \\ \cline{4-9} 
\multicolumn{1}{c}{}                                                                  & \multicolumn{1}{c}{}                        &                           & TextConsis ↑     & BGConsis ↑       & ImageQual ↑      & SubConsist ↑     & ↓ FLOPs                & SpeedUp↑      \\ \hline
\multirow{2}{*}{\begin{tabular}[c]{@{}l@{}}61*448*832\\ 23,296  tokens\end{tabular}}    & Full Attention                              & -                         & 30.51\%          & 93.13\%          & 61.99\%          & 89.43\%          & $1.51 \times 10^{12}$        & -             \\
                                                                                      & \textbf{BSA (Ours)}                          & \textbf{0.93}             & \textbf{30.57\%} & \textbf{93.22\%} & \textbf{62.01\%} & \textbf{89.48\%} &  $1.05 \times 10^{11}$            & \textbf{4.6x} \\ \hline
\multirow{2}{*}{\begin{tabular}[c]{@{}l@{}}157*768*1280\\ 153,600 tokens\end{tabular}} & Full Attention                              & -                         & 31.98\%          & 93.41\%          & 62.97\%          & 90.31\%          & $6.99 \times 10^{13}$        & -             \\
                                                                                      & \textbf{BSA (Ours)}                          & \textbf{0.95}             & \textbf{32.05\%} & \textbf{93.42\%} & \textbf{63.06\%} & \textbf{90.37\%} & $3.49 \times 10^{12}$   & \textbf{6.2x} \\ \hline
\end{tabular}}
\caption{Performance of the training-free BSA.We present a comparison of generation quality and efficiency between BSA and full attention on the Wan2.1-1.3B model without training, focusing on the inference side.} 
	\label{inference}
\end{table*}

\section{Experiment}
\subsection{Dataset and Implement}
\noindent\textbf{Datasets.}
To train the baseline model and BSA from scratch, we selected 300k videos from the Vchitect~\cite{fan2025vchitect} T2V DataVerse and conducted a three-step preprocessing pipeline:
(1) {shot segmentation}; (2) {temporal truncation}; (3) {caption generation}.
We established distinct training protocols for our models to ensure rigorous evaluation. For training the 1.3B model, we conducted ablation studies by processing 300k samples at two resolutions ($448 \times 832$ and $782 \times 1280$) on 8 H100 GPUs, thereby isolating the effects of input size. For finetuning the 14B model, we used the $782 \times 1280$ resolution and trained with 300k samples on 64 GPUs. All experiments maintained consistent datasets and configurations to guarantee fair comparisons.

\noindent\textbf{Metrics.}
We evaluate the generative capability of the models from training efficiency and generation quality. For training efficiency, we measure the acceleration ratio and computation FLOPs. For generative quality, we adopt the five evaluation dimensions of VBench~\cite{huang2024vbench}, specifically: {Text Consistency} for overall coherence, {Dynamic Degree} for motion fidelity, {BG Consistency} for background consistency, {Image Quality} for visual fidelity, and {Sub Consistency} for subject-level consistency.

\noindent\textbf{Baselines.}
In all experiments involving training models from scratch, we compare our sparse attention mechanism against full attention. Since our sparse attention is learnable, we additionally benchmark it against other sparse attention methods~\cite{zhang2025vsa,wu2025vmoba} trained under the same objective.

\noindent\textbf{Implement Details.}
We employ Wan2.1-1.3B~\cite{wan2025wan} as the backbone model for all experiments. For block partitioning, we set $B$ to a size of $(4,4,4)$ with 64 units per dimension. For Query-sparse pruning, the sparsity ratio $r$ is set to 0.5, corresponding to a query block size of 32, and the window size is configured as $(2,2,2)$. 
    \begin{figure}[!h]
\centering
\includegraphics[width=1\linewidth]{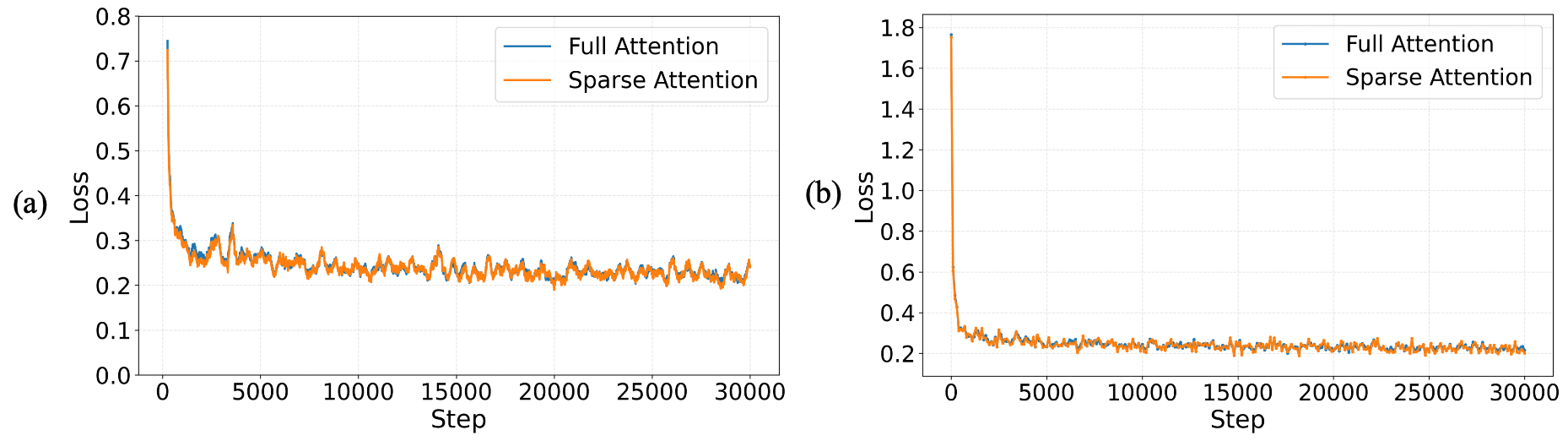}
\caption{Comparison curves of (a) training loss and (b) validation loss for \textit{Sparse Attention} and \textit{Full Attention}.
}
\label{loss}
\end{figure}
Following prior baseline work, we adopt an annealed attention sparsity schedule: training begins with full attention, and every 30 steps, the sparsity is increased by 0.03 until reaching a maximum of 0.9. In Sparse KV, the number of top-$k$ tokens selected is gradually reduced from the total number of blocks to 0.1$\times$ the total, and the dynamic threshold is computed based on the varying $k$. Full training is conducted for 30,000 steps. All experiments are performed on NVIDIA H100 GPUs. Additional implementation details are provided in the Appendix.

\subsection{Quantitative Results}
\textbf{Training-based Comparison.}
We perform all Text-to-video model training based on the Wan2.1-1.3B, training all models to full convergence to ensure fair comparisons.

\noindent\textbf{Loss Comparison.} As shown in Fig.~\ref{loss}, both our proposed Sparse Attention and the Full Attention baseline exhibit stable and smooth pre-training loss curves. Notably, the loss curve of our Sparse Attention model consistently overlaps with that of the Full Attention model, and in most cases, it outperforms Full Attention. Figures~\ref{loss} (a) and (b) illustrate the training and validation loss comparisons, respectively.

\noindent\textbf{Efficiency and Quality Comparison.} As summarized in Table~\ref{table1}, we conduct full scratch training of Sparse Attention and Full Attention on the Wan2.1-1.3B model at two different resolutions: the original resolution ($61 \times 448 \times 832$, 23K tokens) and an extended resolution with longer token sequences ($157 \times 768 \times 1280$, 153K tokens).   Our Sparse Attention demonstrates significant advantages that scale with sequence length. At 23K tokens, it achieves a 12.85x speedup with only 7\% of full attention's FLOPs, while simultaneously improving generative quality on key VBench metrics. These benefits become even more pronounced at 153K tokens, where the speedup reaches 17.79 with just 5\% of the FLOPs, leading to greater gains in consistency due to higher achievable sparsity.

    

\begin{figure}[h]
\centering
\includegraphics[width=0.75\linewidth]{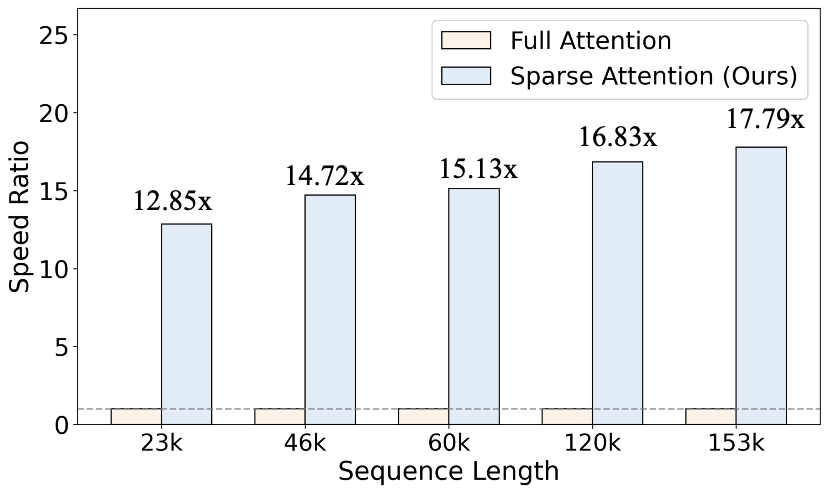}
\caption{{Speedup ratio under varying sequence lengths}.
}
\label{sequence}
\end{figure}

\noindent\textbf{Training on Longer Sequences.}
To assess BSA’s training speedup across different sequence lengths, we tested five lengths (23K, 44K, 59K, 117K, 153K tokens) with consistent settings. As shown in Fig.~\ref{sequence}, speedup increases with sequence length, from 12.85× at 23K to 17.79× at 153K. These results demonstrate that our proposed Sparse Attention becomes increasingly effective at reducing training time as sequence length grows.

    \begin{figure*}[h]
\centering
\includegraphics[width=0.85\linewidth]{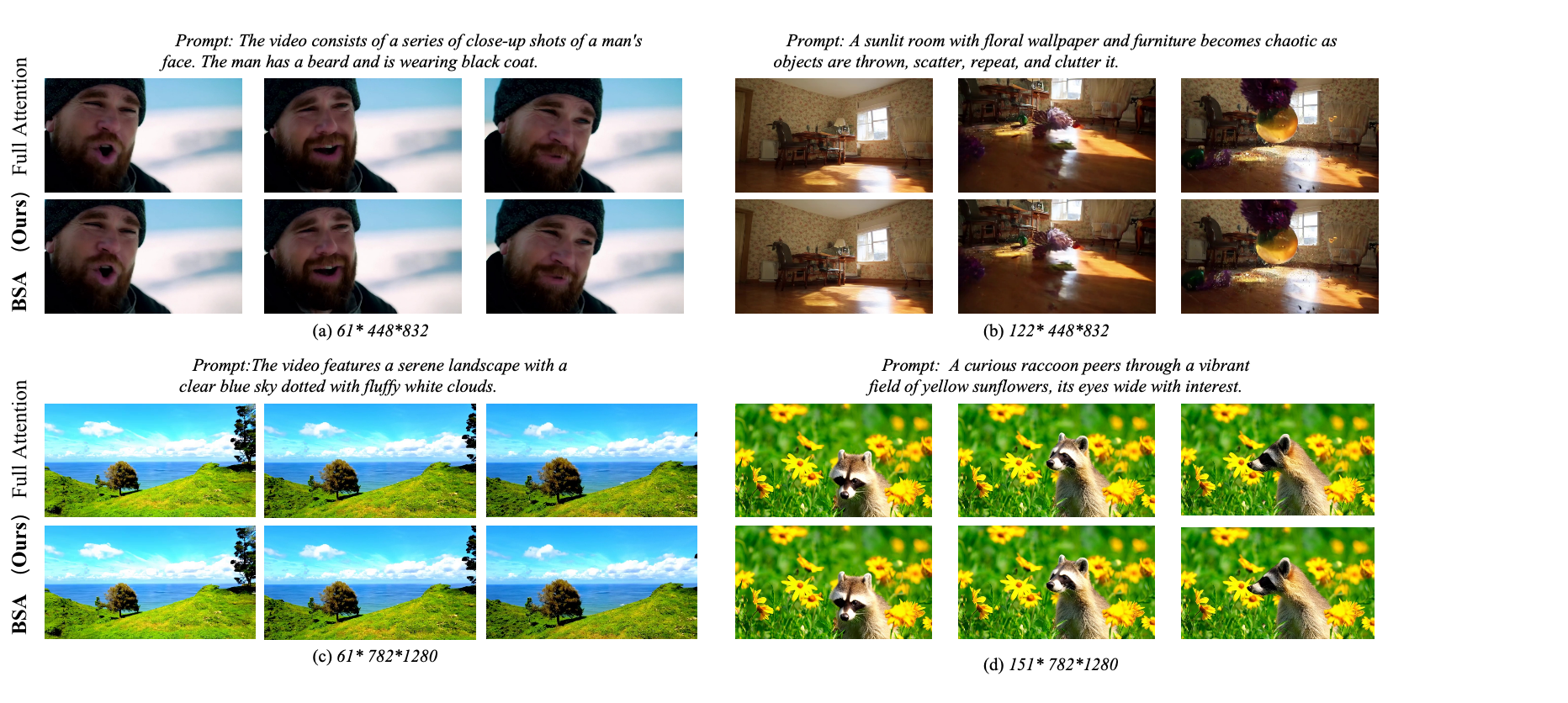}
\caption{Qualitative comparison of text-to-video generation results between full attention and BSA across 4 different sequence lengths.
}
\label{quality}
\end{figure*}
\begin{figure}[h]
\centering
\includegraphics[width=0.75\linewidth]{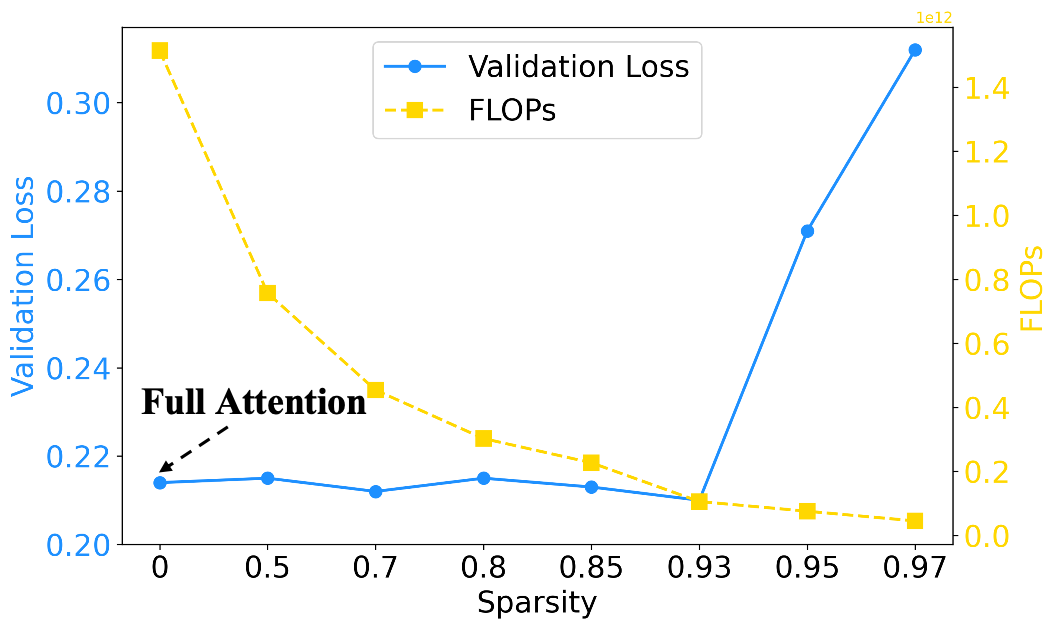}
\caption{{Validation loss and FLOPs under different sparsity.}
}
\label{sparse}
\end{figure}

\noindent\textbf{Sparse Adaptation.}  
To explore how sparsity affects training loss and computational cost, we measured validation loss and FLOPs at various sparsity levels, as shown in Fig.~\ref{sparse}. A sparsity of 0 represents Full Attention training.
Sparsity in our model is controlled by the retained token ratio $r$ for Sparse Query and the dynamic threshold $p$ for Sparse KV, which selects top-$k$ key tokens per attention scores. This creates a trade-off between efficiency and accuracy. 
Fig.~\ref{sparse} shows that as sparsity increases from 0 to 0.93, validation loss remains stable near 0.212, matching Full Attention, while FLOPs drop. Beyond 0.95 sparsity, FLOPs keep decreasing, but validation loss rises sharply, indicating generation quality suffers. The best results are at 0.93 sparsity, where nearly lossless or better generation is achieved with a 13× FLOPs reduction. Notably, optimal sparsity depends on sequence length, longer sequences may benefit from higher sparsity.
Our sparsity computation is more flexible than fixed top-k methods, which require manual tuning for different sequence lengths and model sizes. Instead, our dynamic threshold automatically selects optimal keys based on the data, showing that effective sparse attention should adapt to the input rather than follow a fixed structure. Detailed query sparsity pruning rate ablation experiments are in the Appendix.

\begin{figure}[!h]
\centering
\includegraphics[width=0.7\linewidth]{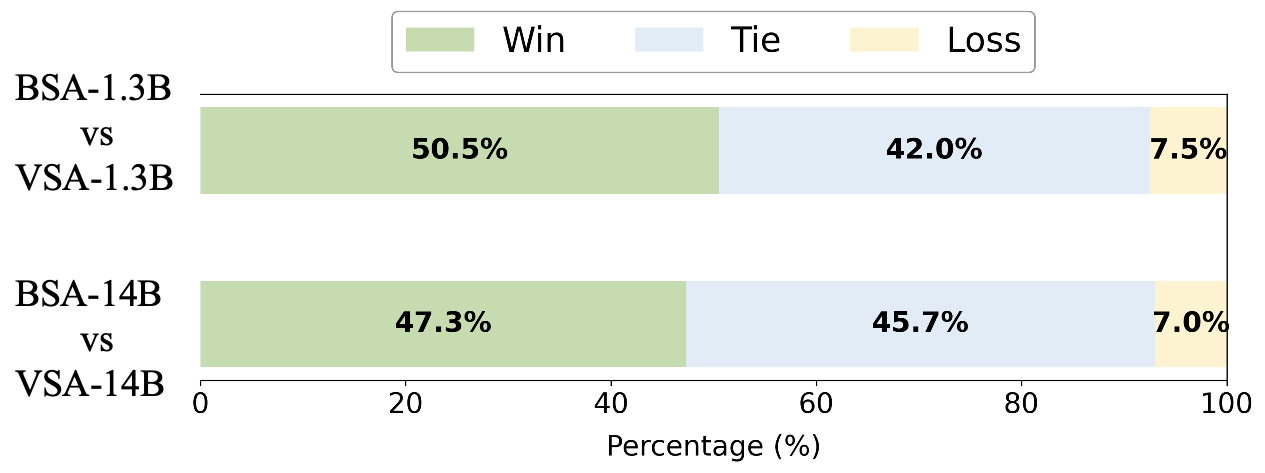}
\caption{{User study between BSA with VSA.}
}
\label{user study}
\end{figure}

\noindent\textbf{User Study.}
To substantiate the performance and scalability of BSA, we benchmarked it against VSA via a human preference study on 1.3B and 14B models, utilizing 210 randomly drawn prompts from MovieGen-bench~\cite{polyak2024movie}. As shown in Figure~\ref{user study}, the study confirms that users consistently favor the outputs from BSA over VSA at both scales. This highlights BSA's capacity to preserve its performance advantage and deliver high-quality results, even when deployed on significantly larger models.

   
\begin{table*}[h]
	\centering
	\scalebox{0.7}{
\small
	\setlength{\tabcolsep}{6pt}
	
\begin{tabular}{lc|c|c|cccc|cc}
\hline
\multicolumn{1}{c}{\multirow{2}{*}{Method}} & \multirow{2}{*}{Settings} & \multirow{2}{*}{Sparsity} & \multicolumn{1}{l|}{\multirow{2}{*}{Validation Loss}} & \multicolumn{4}{c|}{Quality}                                              & \multicolumn{2}{c}{Efficiency}                     \\ \cline{5-10} 
\multicolumn{1}{c}{}                        &                           &                           & \multicolumn{1}{l|}{}                                 & TextConsis ↑     & BGConsis ↑       & ImageQual ↑      & SubConsist ↑     & ↓ FLOPs                     & SpeedUp↑             \\ \hline

\multirow{2}{*}{Sparse Query}               & Original                  & {0.5}              & 0.211                                                 & 32.83\%          & 95.25\%          & 64.34\%          & 92.44\%          & $7.5 \times 10^{11}$  & \textbf{1.96x}            \\
                                            & w/ Window                 & 0.5                       & 0.208                                                 & 32.85\%          & 95.29\%          & 64.36\%          & 92.44\%          & $7.5 \times 10^{11}$  & \multicolumn{1}{c}{\textbf{1.98x}} \\ \hline
\multirow{2}{*}{Sparse KV}                  & Original                  & 0.86                      & 0.210                                                 & 32.84\%          & 95.24\%          & 64.30\%          & 92.41\%          & $2.1 \times 10^{11}$  & \textbf{6.05x}       \\
                                            & w/ Statistic                & 0.89                      & 0.209                                                 & 32.82\%          & 95.25\%          & 64.28\%          & 92.42\%          & $1.67 \times 10^{11}$ & \textbf{6.12x}       \\ \hline
                                            Full Attention                              & -                         & \multicolumn{1}{c|}{0}     & 0.213                                                & 32.71\%          & 95.12\%          & \textbf{64.33\%}          & 92.34\%          & $1.51 \times 10^{12}$                  & -                    \\ \hline
Sparse Query+Sparse KV                      & -                         & \textbf{0.93}             & \textbf{0.212}                                        & \textbf{32.79\%} & \textbf{95.22\%} & {64.29\%} & \textbf{92.39\%} & $1.73 \times 10^{11}$                     & \textbf{12.85x}       \\ \hline
\end{tabular}}
\caption{Ablation Study. All ablation experiments are conducted at a sequence length of 23k to ensure fair comparison. Original (1st row): the selection of a key token in the whole query block rather than in a smaller window block. Window: window-based mechanism in Sparse Query. Original (3rd row): the fixed threshold in Sparse KV. Statistic: statistical dynamic threshold in Sparse KV.} 
	\label{ablation}
\end{table*}

\begin{table*}
	\centering
	\scalebox{0.7}{
\small
	\setlength{\tabcolsep}{10pt}
	
\begin{tabular}{llccccccc}
\hline
\multicolumn{1}{c}{\multirow{2}{*}{Seq\_len}}                                         & \multicolumn{1}{c}{\multirow{2}{*}{Method}} & \multirow{2}{*}{Sparsity} & \multicolumn{4}{c}{Quality}                                               & \multicolumn{2}{c}{Efficiency}          \\ \cline{4-9} 
\multicolumn{1}{c}{}                                                                  & \multicolumn{1}{c}{}                        &                           & TextConsis ↑     & BGConsis ↑       & ImageQual ↑      & SubConsist ↑     & ↓ FLOPs                & SpeedUp↑       \\ \hline
\multirow{3}{*}{\begin{tabular}[c]{@{}l@{}}61*448*832\\ 23,296 tokens\end{tabular}}    & MoBA~\cite{lu2025moba}                                       &           0.80             & 32.56\%          & 95.14\%          & 64.14\%          & 92.05\%               &   $3.02 \times 10^{11}$                      &        1.2x        \\
                                                                                      & VSA~\cite{zhang2025vsa}                                         & 0.87               & 32.65\%          & 95.03\%          & 64.25\%          & 92.21\%          & $1.96 \times 10^{11}$                        & 4.5x            \\ \cline{2-9} 
                                                                                      & \textbf{Sparse Attention (Ours)}            & \textbf{0.93}             & \textbf{32.79\%} & \textbf{95.22\%} & \textbf{64.29\%} & \textbf{92.39\%} & $1.05 \times 10^{11}$             & \textbf{12.85x} \\ \hline
\multirow{3}{*}{\begin{tabular}[c]{@{}l@{}}157*768*1280\\ 153,600  tokens\end{tabular}} & MoBA~\cite{lu2025moba}                                        &   0.80     
    & 34.34\%          & 93.05\%          & 65.34\%          & 93.49\%          &             $2.62 \times 10^{12}$              &          2.3x     \\
                                                                                      & VSA~\cite{zhang2025vsa}                                         & 0.87                      & 34.72\%          & 93.22\%          & 65.87\%          & 93.72\%          & $4.54 \times 10^{11}$                        & 6.2x            \\ \cline{2-9} 
                                                                                      & \textbf{Sparse Attention (Ours)}            & \textbf{0.95}              & \textbf{34.93\%} & \textbf{93.41\%} & \textbf{66.03\%} & \textbf{94.13\%} &   $3.49 \times 10^{12}$  & \textbf{17.79x} \\ \hline
\end{tabular}}
\caption{The comparison of generation quality and efficiency between BSA and the most related works, MoBA~\cite{lu2025moba} and VSA~\cite{zhang2025vsa}.} 
	\label{compare}
\end{table*}

\noindent\textbf{Inference-based Comparison.}
As shown in Table~\ref{inference},  we evaluated BSA for training-free inference, achieving 4.6× and 6.2× speedups on 23K and 153K token sequences. Unlike competing methods, its acceleration is lossless and often outperforms Full Attention. BSA does this by dynamically selecting semantically critical tokens instead of fixed sparsity patterns, thus avoiding performance degradation.

\noindent\textbf{Qualitative Results.} 
As illustrated in Fig.~\ref{quality}, we present qualitative comparisons of T2V generation results across four representative cases, spanning different sequence lengths and resolutions (448$\times$832 and 782$\times$1280). 
For fairness, we sample the same frames from videos generated by full attention and our proposed sparse attention, selecting 3 representative frames. 
Across all examples, our BSA achieves generation quality comparable to full attention without perceptible degradation. 
Sparse attention consistently generates high-quality videos across various scenarios. It produces realistic facial expressions and movements (a), preserves fine details and precise alignment in complex scenes (b), faithfully renders textures in high-res landscapes (c), and maintains fidelity in challenging cases with intricate backgrounds and fur textures (d). More comparisons and generated videos are in the Appendix.

\noindent\textbf{Comparison with Training-based Attentions.} 
As shown in the Table~\ref{compare}, we provide a comparison with the relevant training-based sparse attention methods, such as MoBA~\cite{lu2025moba} and VSA~\cite{zhang2025vsa}. Our BSA achieves an advantage in speedup over others, and also delivers superior generation quality compared to these sparse attention approaches.



\subsection{Ablation Study}
To investigate the impact of Sparse Query and Sparse KV on acceleration and generation quality, we conduct comprehensive ablation experiments, as shown in Table~\ref{ablation}. More ablation studies are in the appendix.  
\subsubsection{Sparse Query}

\textbf{Original Sparse Query.}  
Without KV-sparsity, we first evaluate a baseline Sparse Query. Each block selects a center token and prunes redundant tokens based on similarity. As shown in Row 1, with a pruning rate r=0.5, validation loss is even lower than full attention, indicating lossless training. This achieves 50\% sparsity and 1.96× speedup. Sparse Query also consistently surpasses full attention on all VBench metrics, showing it removes redundant tokens while preserving meaningful queries.

\noindent\textbf{Sparse Query with Window Size Selection.}  
We further extend Sparse Query by adopting multiple center tokens via window partitioning. Specifically, for a block of size $(4,4,4)$, we split it into $(2,2,2)$ windows, yielding 8 sub-blocks. A center token is selected within each window, and pruning is conducted locally. As shown in Row 2 of Table~\ref{ablation}, this approach achieves lower validation loss and better VBench performance at the same sparsity and FLOPs, confirming that window-based selection more effectively preserves meaningful tokens and reduces redundancy.
\subsubsection{Sparse KV}

\textbf{Original Sparse KV.}  
Independent of Sparse Query, we test KV-sparsity with a fixed threshold $p$. As shown in Row 3, this method achieves $0.86$ sparsity, a $6.05\times$ acceleration, and an $8.6\times$ reduction in FLOPs, while maintaining a validation loss comparable to that of full attention. Although a slight drop is observed in the \textit{ImageQual} metric, improvements in text, background, and subject consistency compensate, resulting in near-lossless generation quality overall.

\noindent\textbf{Sparse KV with Statistical Dynamic Threshold.}  
We further enhance KV-sparsity by replacing the fixed $p$ with a dynamic threshold, adaptively computed from the attention score distribution within each block. As shown in Row~4 of Table~\ref{ablation}, this approach achieves higher sparsity and greater acceleration with similar validation loss. It enables KV-sparsity to adjust to block redundancy, overcoming fixed k limitations. By using input-dependent attention scores, it selects more informative KV pairs, resulting in lower validation loss and better generative quality.

\noindent\textbf{Sparse Query + Sparse KV.}
Combining Sparse Query and Sparse KV (Row 6, Table~\ref{ablation}) achieves the best results, with validation loss and generation quality matching or exceeding those of full attention. Their orthogonality allows additive sparsity gains without sacrificing quality, confirming the effectiveness of our design. Additionally, the extra computational overhead is negligible, highlighting the practical efficiency of our sparsification strategies.

\section{Conclusion}
In this work, we presented BSA, a trainable sparse attention framework that jointly sparsifies Queries and Key-Value pairs within 3D full attention. By exploiting the inherent and dynamic sparsity of video sequences, BSA achieves substantial reductions in FLOPs and training time while surpassing the generative quality of full attention. Extensive experiments validate that BSA scales efficiently to video diffusion models, offering a practical solution to computational bottlenecks of DiTs.

{
    \small
    \bibliographystyle{ieeenat_fullname}
    \bibliography{main}
}


\end{document}